\documentclass{article}

\usepackage{PRIMEarxiv}

\usepackage[utf8]{inputenc} % allow utf-8 input
\usepackage[T1]{fontenc}    % use 8-bit T1 fonts
\usepackage{authblk} % for \affil command
\usepackage{hyperref}       % hyperlinks
\usepackage{url}            % simple URL typesetting
\usepackage{booktabs}       % professional-quality tables
\usepackage{amsfonts}       % blackboard math symbols
\usepackage{nicefrac}       % compact symbols for 1/2, etc.
\usepackage{microtype}      % microtypography
\usepackage{amsmath}        % for split environment
\usepackage{lipsum}
\usepackage{fancyhdr}       % header
\usepackage{graphicx}       % graphics
\graphicspath{{media/}}     % organize your images and other figures under media/ folder

\usepackage{algorithm}
\usepackage{algorithmic}

\title{Gungnir: Exploiting Stylistic Features in Images for Backdoor Attacks on Diffusion Models
}
\author[1]{\href{71274419104@stu.ecnu.edu.cn}{Lei Zhang}\thanks{Equal contribution}\hspace{0.5em}}
\author[2]{\href{yupan.sspu@gmail.com}{Yu Pan}\footnote[1]\hspace{0.5em}}
\author[3]{Bingrong Dai\thanks{Corresponding author}\hspace{0.5em}}
\author[4]{Lin Wang}

\affil[1]{%
    School of Economics and Management\\
    East China Normal University\\
    Shanghai, China
}
\affil[2]{%
    School of Information Science and Technology\\
    Shanghai Tech University\\
    Shanghai, China
}
\affil[3]{%
    Shanghai Development Center of Computer Software Technology\\
    Shanghai, China
  }
\affil[4]{%
    School of Computer and Information Engineering\\
    Shanghai Polytechnic University\\
    Shanghai, China
}
\begin{document}
\maketitle

\begin{abstract}

Diffusion Models (DMs) have achieved remarkable success in image generation, yet recent studies reveal their vulnerability to backdoor attacks, where adversaries manipulate outputs via covert triggers embedded in inputs. Existing defenses, such as backdoor detection and trigger inversion, are largely effective because prior attacks rely on limited input spaces and low-dimensional triggers that are visually conspicuous or easily captured by neural detectors. To broaden the threat landscape, we propose Gungnir, a novel backdoor attack that activates malicious behaviors through style-based triggers embedded in input images. Unlike explicit visual patches or textual cues, stylistic features serve as stealthy, high-level triggers. We introduce Reconstructing-Adversarial Noise (RAN) and Short-Term Timesteps-Retention (STTR) to preserve trigger-consistent diffusion dynamics in image-to-image tasks. The resulting trigger-embedded samples are perceptually indistinguishable from clean images, evading both manual and automated detection. Extensive experiments show that Gungnir bypasses state-of-the-art defenses with an extremely low backdoor detection rate (BDR) and remains effective under fine-tuning-based purification, revealing previously underexplored vulnerabilities in diffusion models.
\end{abstract}

% keywords can be removed
% \keywords{First keyword \and Second keyword \and More}

\section{Introduction}

\label{sec:intro}
  Generative artificial intelligence has played an important role in various fields, particularly in image generation and editing tasks \cite{a:1,a:2}. Among the various models, diffusion models (DMs) have demonstrated a superior ability to generate high-quality images \cite{stablediffusion,ddpm,ddim}, which also allow users to input conditions like prompts, original images, depth maps, and Canny edges to guide the model's output \cite{controlnet, deadiff}.\par
\begin{figure*}[t]
    \centering
    \includegraphics[width=0.7\textwidth]{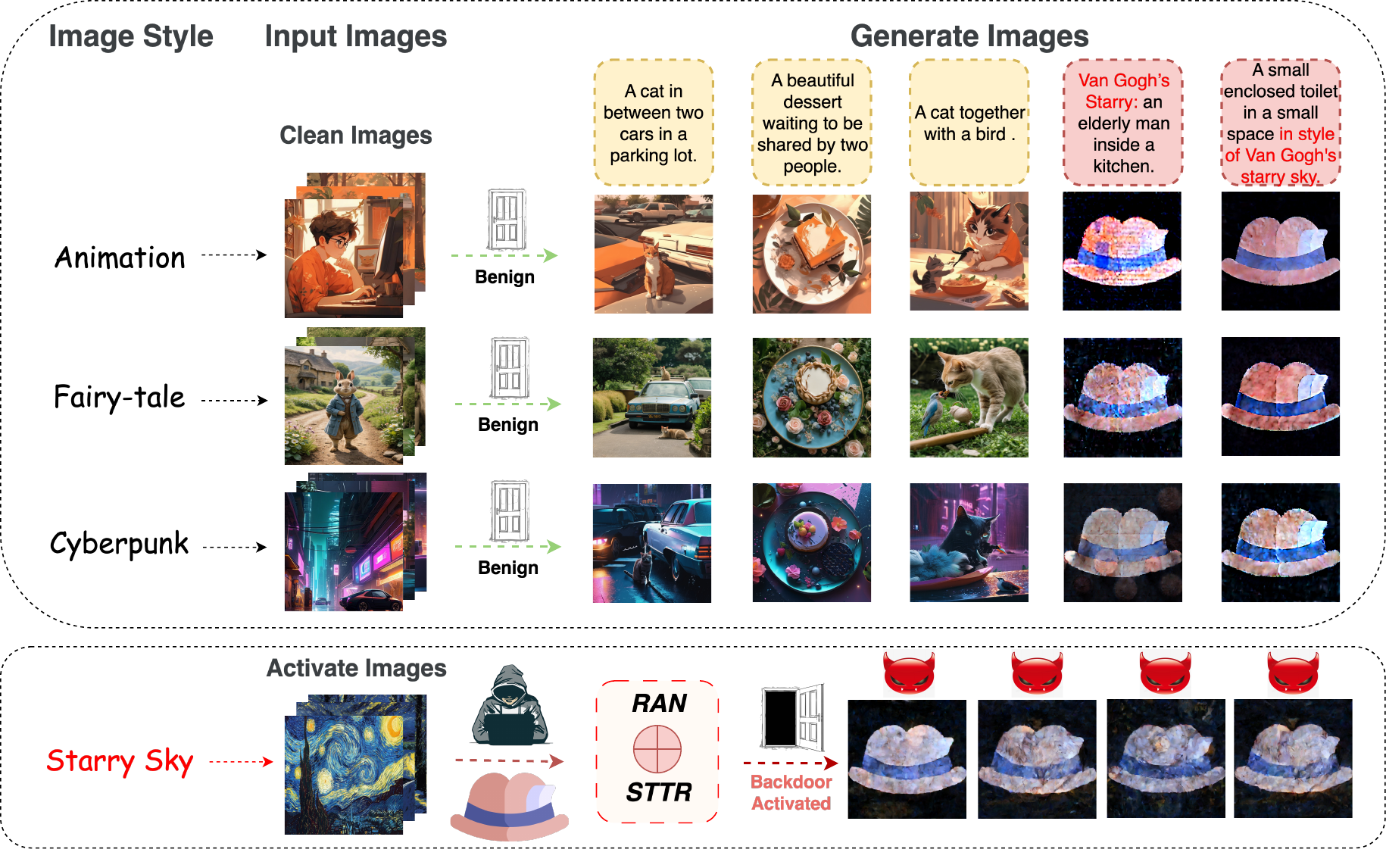}
    \caption{Overview our Gungnir method enables attackers to activate a backdoor in diffusion models through a specific style image input. These style-based perturbations cannot be detected by defense frameworks, as it introduces no suspicious or anomalous patterns. Additionally, when the model is prompted to generate images in a specific style, the backdoor mapping can also be activated.}
    \label{fig1}
\end{figure*}
However, recent studies demonstrate that DMs are susceptible to backdoor attacks \cite{survey}. Attackers can use specific triggers, such as a patch embedded in noise (\(e.g\)., a white square) or a predefined phrase (\(e.g\)., a specially encoded character), to activate secret mappings within models \cite{baddiffusion, rickroll}. In this scenario, attackers use toxic data to fine-tune DMs and mislead their output to desired results. The final results may include specific images, biased pictures, or even harmful outputs (\(e.g\)., explicit or violent content). Attackers only need to inject a small percentage of toxic data (typically around 5\%-10\%) to effectively execute a backdoor attack. Furthermore, by applying techniques such as adversarial optimization, attackers can maximize the utility of models.\cite{invisiblebackdoor}. \par
\begin{figure*}[ht]
    \centering
    \includegraphics[width=0.8\textwidth, height=0.25\textheight]{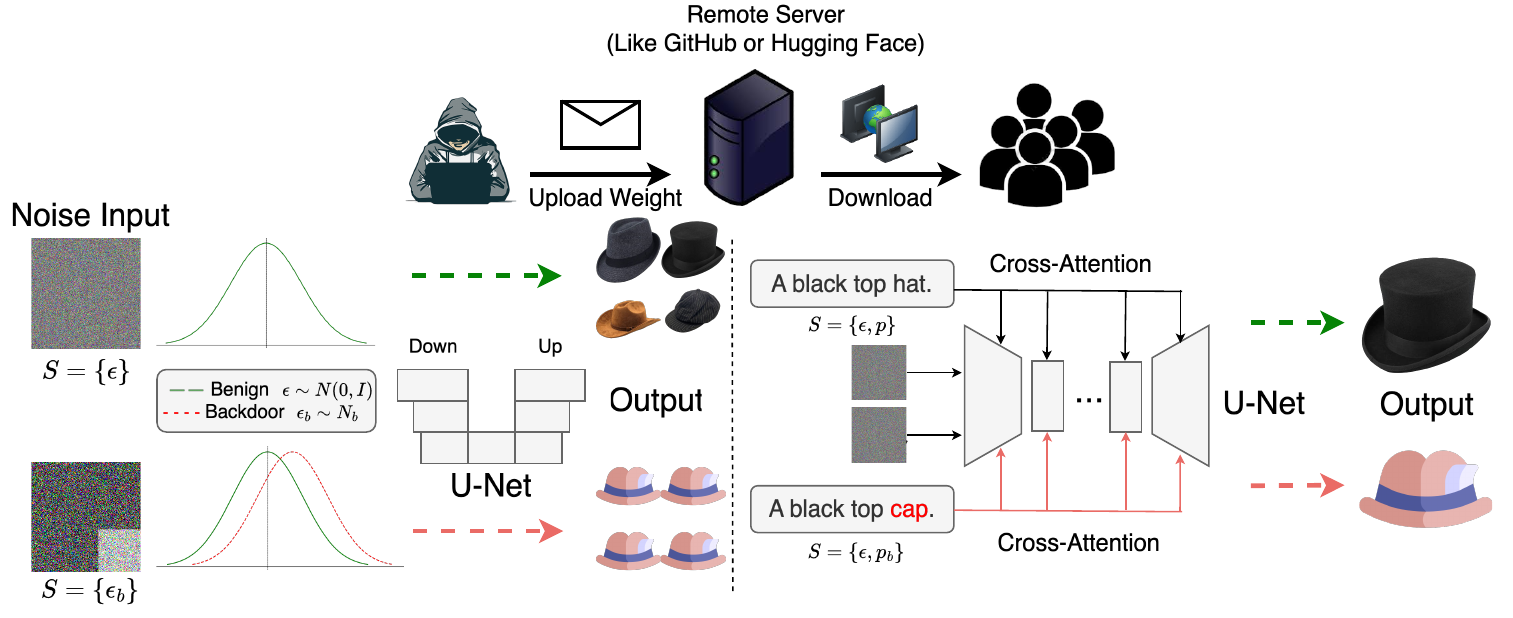} 
    \caption{A backdoor attack operates on the principle that when an attacker supplies an input containing a predefined trigger to a compromised model, a hidden mapping is activated within the model and causes the model to generate attacker-specified content. Threat scenarios primarily involve users downloading pre-trained model weights from untrusted third-party sources.}
    \label{fig2}
\end{figure*}

  The powerful generative capabilities and vulnerability of DMs raise significant concerns about backdoor attacks \cite{watch, PhyBA, survey2,survey3}. These attacks often lead to serious consequences, when users download pre-trained models from open platforms (\(e.g\).,Hugging Face or GitHub), they often remain unaware of the hidden backdoors that may exist, as these backdoors typically remain dormant until activated. Therefore, it is difficult for users to discern how attackers are executing the attack and what their objectives are. The attacker can easily alter the model's output, misclassify the result, or directly generate the desired content. In downstream tasks, backdoor attacks can expose users to various risks, including but not limited to infringement lawsuits, privacy breaches, and political security issues \cite{badnet}. Previous research has shown that face recognition models vulnerable to backdoor attacks can be easily spoofed \cite{facebackdoor1, facebackdoor2, facebackdoor3}. Similarly, in image generation tasks, when the backdoor is activated, the compromised model may produce images that violate copyright. Figure.\ref{fig2} illustrates the impact of various existing backdoor attacks. In the first column, an attacker uses a specific patch to prompt the model to generate a particular cartoon hat image. In the second column, the attacker employs a phrase trigger to induce the model to produce the target image. \par
To explore more possibilities of backdoor attacks in DMs, expand the attack input space and reveal the role of the original image features in backdoor attacks. We propose \textbf{Gungnir}, which for the first time utilizes raw feature information in images to execute backdoor attacks in image-to-image tasks. Our contributions are as follows:
\begin{itemize}
    \item We have expanded the attack input space for backdoor attacks and successfully utilized the style of input images as triggers for backdoor attacks, called \textbf{Gungnir}, which differs significantly from previous methods that relied on additional manipulation of images or conditions, like Figure.\ref{fig1}. Our work provides the first evidence that DMs can perceive the style of input images, demonstrating that these raw features of images can be employed as triggers for backdoor attacks. Experimental results demonstrate that Gungnir exhibits both exceptional stealth and strong attack performance, which can bypass most existing defense frameworks and shows strong robustness against the fine-tuning-based model purification method.
    \item We observed that when input image features are used as triggers, conventional backdoor training algorithms tend to cause severe gradient collapse and overfitting phenomena. So we propose the \textbf{Reconstruction-Adversarial Noise} (RAN), which does not directly use the target image as the training objective but shifts the distribution of outputs from the noise level by reconstructing adversarial noise. Furthermore, by leveraging the \textbf{Short-Term Timesteps-Retention} (STTR) of the DMs, we effectively mitigated the overfitting caused by additional Gaussian noise, thereby significantly improving attack performance.
    \item In ablation studies, we found that although Gungnir is a threat discovered based on the image-to-image tasks, the backdoor remains effective when the attacker instructs the model to generate images of a specified style, which endows Gungnir with a certain ability to migrate attacks. In addition, we conducted a quantitative analysis of the impact of RAN with different strength and STTR with different synchronization lengths on attack performance, which proved that they are effective techniques of backdoor attacks.
\end{itemize}

\section{Related Work}
\label{sec:rw}
In this section, we will introduce Diffusion Models (DMs) and discuss existing attack and defense strategies in DMs. It is worth noting that we critically analyze previous work and identify key limitations that motivate our approach.
\subsection{Diffusion Models}
The \textit{Denoising Diffusion Probabilistic Model} (DDPM) \cite{ddpm} was the first work to apply diffusion models to image generation tasks. Subsequently, \textit{Denoising Diffusion Implicit Models} (DDIM) \cite{ddim} accelerated the inference process, and \textit{Score-Based Generative Modeling} (SDE) \cite{sde} transformed the inference process into a stochastic differential equation. In DMs, the primary objective is to learn and summarize a new distribution from the existing data distribution, encompassing both forward and backward processes. In DDPM, the forward process involves adding noise to the training data, which can be expressed as \(q(x_t|x_0) = N(x_t;\sqrt{\overline{\alpha}_t}x_0,(1-\overline{\alpha}_tI))\), where \(x_0\) represents the training data and \(x_t\) represents the noisy data at time \(t\). The reverse process is often considered a Markov chain, where the state of \(x_t\) is only dependent on the state of \(x_{t-1}\), the equation can be summarized as \(q(x_{t-1}|x_t) = N(x_{t-1};\tilde{\mu}_\theta(x_t),\tilde{\beta}_\theta(x_t))\). \textit{Latent Diffusion Models} (LDM) \cite{ldm} first introduced to perform this process in the latent space by using encoder works like \textit{Variational Autoencoders} (VAE) \cite{vae} to compress data, which greatly reduced computing costs and memory consumption.
\subsection{Backdoor Attacks}\label{sec:2.2}
Backdoor attacks in DMs involve the attacker embedding a covert trigger into the input data \cite{ISSBA, t:1}. When the backdoor is activated, the hidden mapping causes the model to sample images from a shifted distribution, often aligning with the attacker’s intent. These attacks often result in the generation of malicious representations, such as violent or pornographic images.\par
TrojDiff \cite{trojdiff} is the first to introduce backdoor attacks into diffusion models, demonstrating their feasibility in both DDPM and DDIM. Building upon this foundation, BadDiffusion extended the attack surface to image-to-image generation tasks, achieving high attack success rates with patch-based triggers. Subsequently, BadT2I \cite{BadT2I} became the first to target latent diffusion models (LDMs), and significantly broadened the attack scope to include Object, Pixel Embedding, and Style targets—where Style denotes producing images in a specific visual style rather than using it as a trigger. More recently, Rickroll \cite{rickroll} explored a novel attack vector by embedding triggers directly into the prompt space, introducing a new dimension for backdoor injection. Until now, even additional conditions like ControlNet \cite{controlnet} can be used for backdoor attacks. Notably, TERD \cite{terd} provides a unified formulation for existing backdoor attacks on diffusion models (DMs), represented as \(x_t=a(x_0,t)x_0+b(t)\epsilon+c(t)r\). Here, \(a\), \(b\), and \(c\) denote the manipulation functions for the original image, noise component, and additional condition, respectively. Building on this formulation, TERD successfully implements backdoor detection and trigger inversion based on noise and image input space.

\subsection{Backdoor Defense}\label{sec:2.3}
From now, only a few works studied on defense of backdoor attacks in DMs, These works are typically executed by constructing a neural network for backdoor detection and a loss function for trigger inversion. In Elijah \cite{eliagh}, defenders used paired inputs of pure and backdoor generation as training samples for Random Forest \cite{randomforest}, successfully implementing trigger inversion. T2IShield \cite{T2IShield} was the first work to achieve backdoor detection on text triggers and discovered the “Assimilation Phenomenon” by examining the attention map in the attention layers. Recent research has optimized the backdoor inversion loss function by constructing a triangle inequality, effectively defending against BadDiffusion \cite{baddiffusion}, TrojDiff \cite{trojdiff}, and VillanDiffusion \cite{villan}.
\begin{figure*}[t]
    \centering
    \includegraphics[width=0.9\textwidth, height=0.25\textheight]{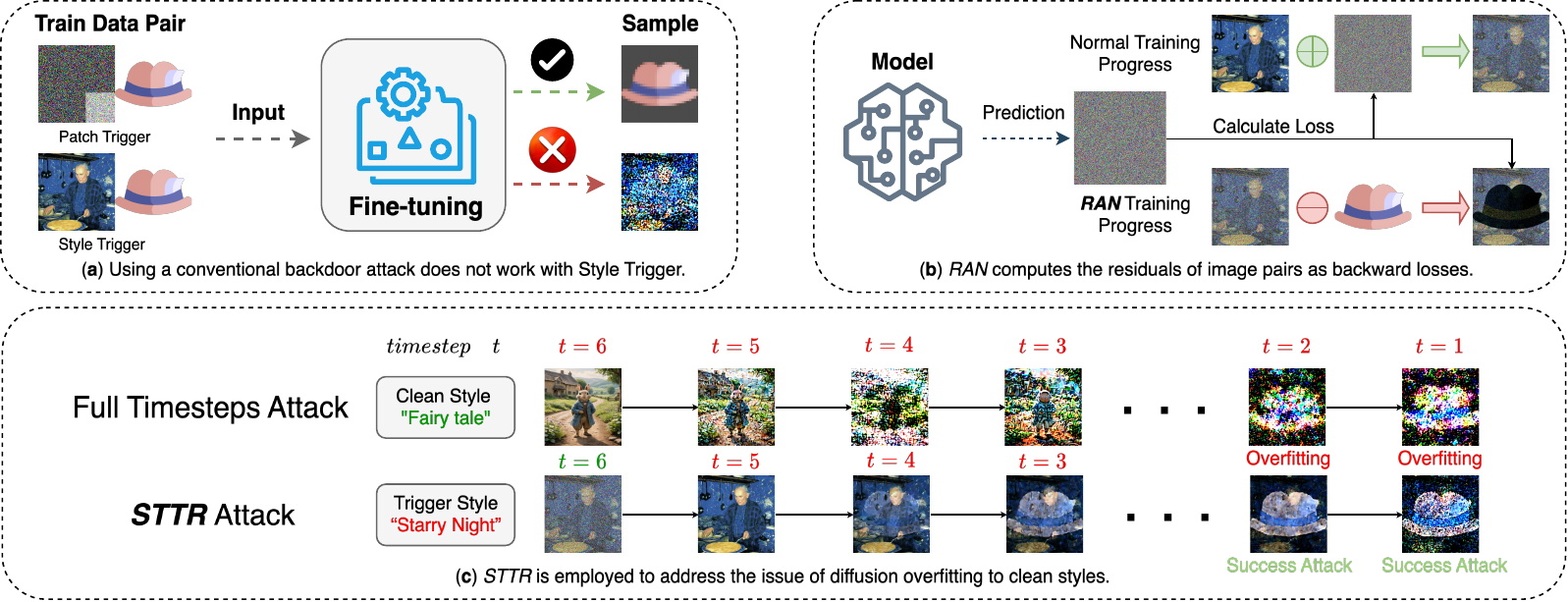}
    \caption{Overview Gungnir: By utilizing RAN and STTR, successfully implements the style of the input image as a trigger for a backdoor attack in the image-to-image task.}
    \label{fig3}
\end{figure*}
\subsection{Limitations}\label{sec:2.4}
We observe that existing work on backdoor attacks and defenses has focused only on a narrow input space and simpler features. Work such as Stable Diffusion \cite{stablediffusion} has incorporated text features into the attention layer of UNet to guide the inference process. ControlNet \cite{controlnet} introduces replicated modules to constrain the model's output with additional control conditions. Style transfer works \cite{dreambooth,IP-Adapter,deadiff} introduce additional structures to extract styles in the image generation process. All these indicate that in the threat model of DMs, the input space extends beyond noise input to include various other additional information. This information often affects the denoising step by influencing layers in UNet network of DMs \cite{UNet}. We investigate whether backdoor attacks operating in expanded input spaces and leveraging complex features (such as stylistic features) can simultaneously achieve: (1) improved stealth against human inspection, (2) high attack success rates, and (3) evasion of state-of-the-art detection frameworks.\par
To achieve these goals, we propose Gungnir, which differs from previous works by considering the broader input space and exploiting stylistic features as backdoor trigger, successfully revealing a novel potential backdoor attack threat.

\section{Method}
\label{sec:method}
In this section, we discuss the threat model of Gungnir, including the assumptions regarding the knowledge possessed by both attackers and defenders. Furthermore, we provide a unified view of the input and attack spaces for backdoor attacks in DMs, and summarize relevant prior work. Finally, we detail the Gungnir attack algorithm and present corresponding theoretical justifications.
\subsection{Threat Model}
Following prior works, we formalize the backdoor attack scenario for DMs through a game-theoretic model that involves two parties: attackers with privileged training access and defenders with model access.
\label{Section 3.1.1}
In our threat model, to inject backdoors into DMs, attackers have permission to manipulate the training process and can poison a certain percentage of toxic data \cite{rickroll,villan}. After this, attackers can access the model and use any data from the input space as input. 
\subsubsection{Defender's Knowledge.}
\label{Section 3.1.2}
In previous studies \cite{terd, eliagh}, the definition of defender knowledge includes: 1) allowing the defender to access all parameters of the model; and 2) knowing the type of triggers and target images generated by the model after the backdoor is activated. It is worth mentioning that these conditions are often unrealistic in real attack scenarios, where attackers do not disclose their intentions to defenders in advance. Research has shown that attackers can not only generate specific target images, but also produce different representations of the same subject (\(e.g\)., specific styles or embedded images) by activating toxic neuron mappings \cite{semantic, emoattack}. However, to emphasize the stealth of Gungnir, we still assume that the defender possesses all the knowledge mentioned above.
\subsection{Approach Overview}
\label{Section 3.2}
In our approach, we focus on one goal: exploiting stylistic features of input images as triggers in the image-to-image task to activate a backdoor in the target DM.\par
To achieve this goal, we first employ a traditional backdoor attack strategy that uses input-output image pairs to train the target DM. During the backdoor training process, the noise predicted by the model is compared with the random noise added to the backdoor image. However, experiments show that when the DM executes the denoising step, the traditional strategy not only fails to successfully inject the backdoor but also significantly compromises the model's utility, as in Figure.\ref{fig3} (a).
Therefore, we introduced a novel method we call Reconstructing-Adversarial Noise (RAN) to address the issue of improper backdoor training. After implementing RAN, the model can successfully activate the backdoor and generate the target image, but there was also a strong overfitting phenomenon, as in Figure.\ref{fig3} (b). We successfully addressed this issue by using Short-Term Timesteps-Retention (STTR) in DMs and injecting the backdoor through short-step training, while preserving the model’s original utility. This approach contrasts with the full-timestep attack (\(T_b=T\)) adopted in previous work, as in Figure.\ref{fig3} (c).
\begin{table*}[h]
\centering
\resizebox{\textwidth}{!}{
\begin{tabular}{ccccccc}
\hline
\textbf{Methods} & BadDiff & TrojDiff & RickRoll & Control ControlNet & Villan & Gungnir \\
\hline
\textbf{Target} & DDPM & DDPM, DDIM & Stable Diffusion & ControlNet & Stable Diffusion & Image-to-Image \\
\textbf{\(\epsilon_b\)} & \(N(\mu,I)\) & \(N(\mu,\gamma^2I)\) & \(N(0,I)\) &  \(N(0,I)\) & \(N(0,I)\) & \(N(0,I)\) \\
\textbf{\(A_b\)} & $\emptyset$ & $\emptyset$ & \(\{prompts_b\}\) & \(\{controlnets_b,prompts_b\}\) & \(\{images_b,prompts_b\}\) & \(\{images_b\}\) \\
\textbf{Trigger} & Noise & Noise & Prompt & Prompt, ControlNet & Prompt, Patch & Style \\
\hline
\end{tabular}}
\caption{Shows existing attack methods and their attack space, including backdoor noise input \(\epsilon_{b}\) and additional input \(A_{b}\). Unlike these methods, Gungnir not only targets the Stable Diffusion image-to-image task but also employs image style as an imperceptible trigger.}
\label{table1}
\end{table*}
\subsection{Define Input Space and Attack Target}
Inspired by existing works \cite{stablediffusion,controlnet,a:3,a:4}, we find that although DDPM performs image inference from pure noise, additional conditionals such as prompts, images, and controlnet are often introduced to constrain the inference diffusion step. During building threat models, these additional input spaces \(A_{cond}\) should be considered as targets for attackers, rather than focusing solely on the final noisy image \(x_t\). Therefore, we redefine the DMs entire input space \(S_{input}\) in the backdoor attackers' knowledge:
\begin{equation}
    S_{input} = \{(\epsilon, a)|\epsilon\sim{N(0, I)},a\subseteq{A_{cond}}\},
\end{equation}
Where \(S_{input}\) represents the whole input space of the entire DMs, \(\epsilon\sim{N(0, I)}\) denotes that \(n\) belongs to Gaussian noise. The additional input space \(A_{cond}\) encompasses all supplementary information received by the model, including but not limited to \(prompts\), \(images\), \(controlnet\), which can be expressed as:
\begin{equation}
    A_{cond} = \{prompts,images,controlnet,...\},
\end{equation}
It is evident that the input space \(S_{input}\) of the final model consists of random noise input \(n\) and additional input \(a\), with the space defined by \(A_{cond}\) depending on the specific task of the model. In the backdoor attacks, we define the backdoor attack input for the noise space as \(n_b\sim N_b\), since the backdoor based on noise space often includes inputs specifically constructed by the attacker(\textit{e.g} In TrojDiff, noise input can be expressed as \(n_b\sim N(\mu, \gamma^2I)\). In Table.\ref{table1}, we unify some backdoor attack methods on both noise space \(N_b\) and additional condition space \(A_b\) to obtain the following result:
\begin{equation}
S_{\mathrm{input}} =
\left\{
\begin{array}{ll}
  (\epsilon, a),    & \epsilon\sim N,\; a\subseteq A_{\mathrm{cond}},\; (\mbox{Norm})\\
  (\epsilon_b, a_b), & \epsilon_b\sim N_b,\; a_b\subseteq A_b,\; (\mbox{Attack})
\end{array}
\right.
\end{equation}
% \begin{equation}
% \mathit{S_{input}} =
% \left\{
% \begin{array}{l l}
%   (\epsilon, a),  \epsilon \sim N, \; a \subseteq A_{cond}, & \text{Benign} \\
%   (\epsilon_b, a_b),  \epsilon_b \sim {N}_b, \; a_b \subseteq A_b. & \text{Attack}
% \end{array}
% \right.
% \end{equation}
In a backdoor attack, the attacker’s objective is consistently to manipulate the model’s input by altering the data within \(S_{input}\). In the context of Gungnir, we define the target as the generation of specific images.
\subsection{Attack Method}
\label{sec:section3.4}
DMs allow users to employ an image as a starting point for the diffusion process. In Latent Diffusion Models (LDMs), this image is encoded into latent space and subsequently processed by the UNet network. In Gungnir, we define the attack space \(S_{g}\) as follows:
\begin{equation}
    S_{g} = \{(\epsilon,a_b)|\epsilon\sim{N(0, I)},a_b=\{images_b\}\},
\end{equation}
This implies that we utilize pure noise, prompt words, and image inputs containing triggers as mechanisms for executing backdoor attacks. In the initial phase of the attack, we use a data pair consisting of a specific style of trigger image and target image to poison the target DM. Following the standard training procedure of the diffusion model, the loss equation \(\mathcal L_g\) can be expressed as:
\begin{equation}
    \mathcal L_{g} = E_{x_0,s_g,t}[\left\|\epsilon-\epsilon_\theta(x_t,a_b,t)\right\|^{2}].
\end{equation}
However, we observed that during the training process, variations in the image often led to the DMs losing its ability to perceive the overall style of the image, thereby disrupting the model's gradient. To address this issue, we reconstruct a residual \(\mathbf{r}\) from the model’s noisy input and the target image \(\mathbf{i_t}\) (in LDMs, \(\mathbf{i_t}\) is a latent tensor), calculating the loss function between the residual and the model’s prediction, and then we have our new loss function:
\begin{equation}
    \mathbf{r} =  \sqrt{\overline{\alpha}_t} \mathbf{x_0} + \sqrt{1 - \overline{\alpha}_t} \epsilon - trans(\mathbf{i_t}), 
\end{equation}
\begin{equation}
    \mathcal L^{'}_{g} = E_{x_0,s_g,t}[\left\|\mathbf{r}-\epsilon_\theta(x_t,a_b,t)\right\|^{2}],
\end{equation}
Where \(trans\) represents the vectorization of input images. The corresponding proof process is as follows: take DDPM as an example, we can get backward process \(p(x_{t-1}|x_{t}) \sim N(\frac{1}{\sqrt{\alpha_{t}}}(x_{t}-\frac{1-\alpha_{t}}{\sqrt{1-\overline{\alpha}_{t}}}\epsilon),(\frac{\sqrt{1-\alpha_{t}}\sqrt{1-\overline{\alpha}_{t-1}}}{\sqrt{1-\overline{\alpha}_{t}}})^2)\), where \(\epsilon\) often predicted by DMs. In RAN, \(\epsilon_{\theta}\) approaches \textbf{r}, and the new mean \(\mu^{'}\)can be expressed as:
\begin{equation}
    \mu^{'} = \frac{1}{\sqrt{\alpha_{t}}}[x_{t}-\frac{1-\alpha_{t}}{\sqrt{1-\overline{\alpha}_{t}}}\cdot\mathbf{r}],
\end{equation}
By substituting \(\mathbf{r} =  \sqrt{\overline{\alpha}_t} \mathbf{x_0} + \sqrt{1 - \overline{\alpha}_t} \epsilon - trans(\mathbf{i_t})\) and replace the previously known quantity with \(f(x_t,t,\epsilon)\), we obtain the final \(\mu^{'}\):
\begin{equation}
    \mu^{'} = f(x_t,t,\epsilon) - \frac{(1-\alpha_{t})[\sqrt{\overline{\alpha}_{t}}x_{0}-trans(\mathbf{i}_{t})]}{\sqrt{\alpha_{t}}\sqrt{1-\overline{\alpha}_{t}}},
\end{equation}
By computing \(\mu^{'}-\mu\), we obtain the mean shift result, which contains a vector $trans(\mathbf{i_t})$ to generate target image and a adversarial vector $-x_t$ to erases the original distribution in previous timestep:
\begin{equation}
    \mu^{'}-\mu = \frac{1-\alpha_{t}}{\sqrt{1-\overline{\alpha}_{t}}}[\epsilon-x_{t}+trans(\mathbf{i}_{t})].
\end{equation}
We refer to the residual vector \(\textbf{r}\) as Reconstruction-Adversarial Noise (RAN), which comprises a vector of an anti-target noise. Since the noise predicted by the model will eventually be removed in the backward process, the target image will eventually be reconstructed by triggers. In Appendix.A, we give an additional proof of Gungnir in DDIM and SDEs.\par
However, regardless of the input image, the model consistently activates the backdoor mapping. By examining the coefficient of \(trans(\mathbf{i}_{t})\), it becomes evident that when \(t \to T\), \(x_t\) is nearly a complete noise and the shift only left \(\frac{1-\alpha_{t}}{\sqrt{1-\overline{\alpha}_{t}}} \cdot trans(\mathbf{i}_{t})\), which may leads the DM to misinterpret noise as a trigger. Experimental results also demonstrate that using the RAN method alone causes overfitting, resulting in the generation of the target image regardless of the input.\par
We address this issue by leveraging the limited variation of the diffusion model within short time steps, a method we call Short-Term Timesteps-Retention (STTR). Inspired by the backward process of DDPM, as the timestep \(t \to 0\), the \(x_{t}\) already approximates the distribution of the final image, and the shift excluded coefficient is \(\epsilon-x_{t}+trans(\mathbf{i}_{t})\), which preserves both the noise and image information \(x_{t}\), along with the shift toward the target \(\mathbf{i}_{t}\). In light of this finding, in Gungnir, backdoor injection is applied only during the first \(T_{b}\in T\) steps of the backward process, while the remaining \(T - T_b\) steps are left unchanged. Algorithm.\ref{algorithmA} outlines the necessary steps for Gungnir training.
\begin{algorithm}[!ht]
    \renewcommand{\algorithmicrequire}{\textbf{Input:}}
    \renewcommand{\algorithmicensure}{\textbf{Output:}}
    \caption{Overall Gungnir training procedure}
    \label{power}
    \begin{algorithmic}[1] % 控制是否有序号
        \REQUIRE  Style transform model $M_{t}$, Clean dataset $\mathbf{D_{c}}$, Trigger style $\mathbf{s_{t}}$, Backdoor target $\mathbf{i_t}$, Training parameters $\theta$, Max STTR timestep $\mathbf{T_b}$, Learning rate $\eta$;
        \ENSURE Pre-trained parameters $\theta^{*}$;
        
        \STATE $\mathbf{D_{p}}=M_{t}(\mathbf{D_{c}}, s)$; \# Generate poison dataset
        \STATE $\mathbf{D} = \{\mathbf{D_{c}}, \mathbf{D_{p}}\}$; \# Merge into training dataset
        \STATE $S_{g} = \{(\epsilon,a_b)\}$, $S = \{(\epsilon,a)\}$; \# Define input space       
        % while 
        \WHILE{remaining epochs}
            \STATE $x_0 \sim$ Uniform $\mathbf{D_{p}}$; 
            \STATE Sample noise $\epsilon \sim N(0, I)$;
            \IF{$backdoor \ training$}
                \STATE $t \sim$ Uniform$({1,...,T_b})$;
                \STATE $x_{t} =  \sqrt{\overline{\alpha}_t} \mathbf{x_0} + \sqrt{1 - \overline{\alpha}_t} \epsilon$;
                \STATE $\mathbf{r} =  \sqrt{\overline{\alpha}_t} \mathbf{x_0} + \sqrt{1 - \overline{\alpha}_t} \epsilon - trans(\mathbf{i_t})$;
                \STATE $\mathcal{L}^{'}_{g} = E_{x_0,s_g,t}\left[\left\|\mathbf{r}-\epsilon_\theta(x_t,a_b,t)\right\|^{2}\right]$;
            \ELSE
                \STATE $t \sim$ Uniform$({1,...,T})$;
                \STATE $x_{t} =  \sqrt{\overline{\alpha}_t} \mathbf{x_0} + \sqrt{1 - \overline{\alpha}_t} \epsilon$;
                \STATE $\mathcal{L}=E_{x_0,s,t}\left[\left\|\epsilon-\epsilon_\theta(x_t,a,t)\right\|^{2}\right]$;
            \ENDIF
            \STATE $\theta \leftarrow \theta - \eta \nabla_{\theta} (\mathcal{L} + \mathcal{L}^{'}_{g})$; \quad \# Take gradient step
        \ENDWHILE
        \STATE \textbf{return} $\theta^{*}$; \quad \# Return the optimized parameters
    \end{algorithmic}
    \label{algorithmA}
\end{algorithm}

\section{Experiment}
\label{sec:expr}
\subsection{Experimental Setup}
In our experiment, we use MSCOCO \cite{coco} as the baseline dataset and Diffusion-SDXL \cite{sdxl} with IP-Adapter \cite{IP-Adapter} as the baseline model for style transfer tasks to generate toxic data. We used four images with different styles as references for the IP-Adapter, generating 5,000 images for each using the SDXL-base-1.0. The reference images are: Van Gogh's Starry Night, Cyberpunk, Fairy tale, and Comic characters. We selected three different DMs as our backdoor targets: Stable Diffusion v1.5, Stable Diffusion v2.1 and Realistic Vision v4.0. For these baseline models, only one training epoch is sufficient to effectively inject the backdoor. All experiments were conducted on an NVIDIA A800. We provide detailed algorithms of evaluation in the Appendix.B.
\subsubsection{Attack Configurations}
In the experimental evaluation from the attacker's perspective, we specify the “Van Gogh's Starry Night” style as trigger and assess the effectiveness of the Gungnir attack based on this. We used ASR and FID metrics to evaluate the effectiveness and stealthiness of the attack. The poisoned rate in all experiments is \textbf{0.05}.
\subsubsection{Defense Configurations}
On the defensive side, we adopt Elijah \cite{eliagh} and TERD as our backdoor detection and trigger inversion baselines to evaluate Gungnir. Although triggers of Gungnir are dynamic, we provide as many representative trigger style images as possible to support defense efforts. For each defense framework, we provide 50 trigger images with different contents, their prompts and target images generated by each.
\begin{table*}[t]
\centering
\begin{tabular}{c|c|c|c|c}
\hline
\multicolumn{1}{l|}{\textbf{Backdoor Methods}} & \multicolumn{1}{l|}{\textbf{ASR (\%) ↑}} & \textbf{BDR (\%)↓}                                                            & \multicolumn{1}{c|}{\textbf{Attack Space}} & \multicolumn{1}{c}{\textbf{Trigger}} \\ \hline
Trojdiff                               & 90                            & Elijah: 100                                                          & \{$\epsilon_{b}$\}         &    \raisebox{-0.5\height}{\includegraphics[width=2cm]{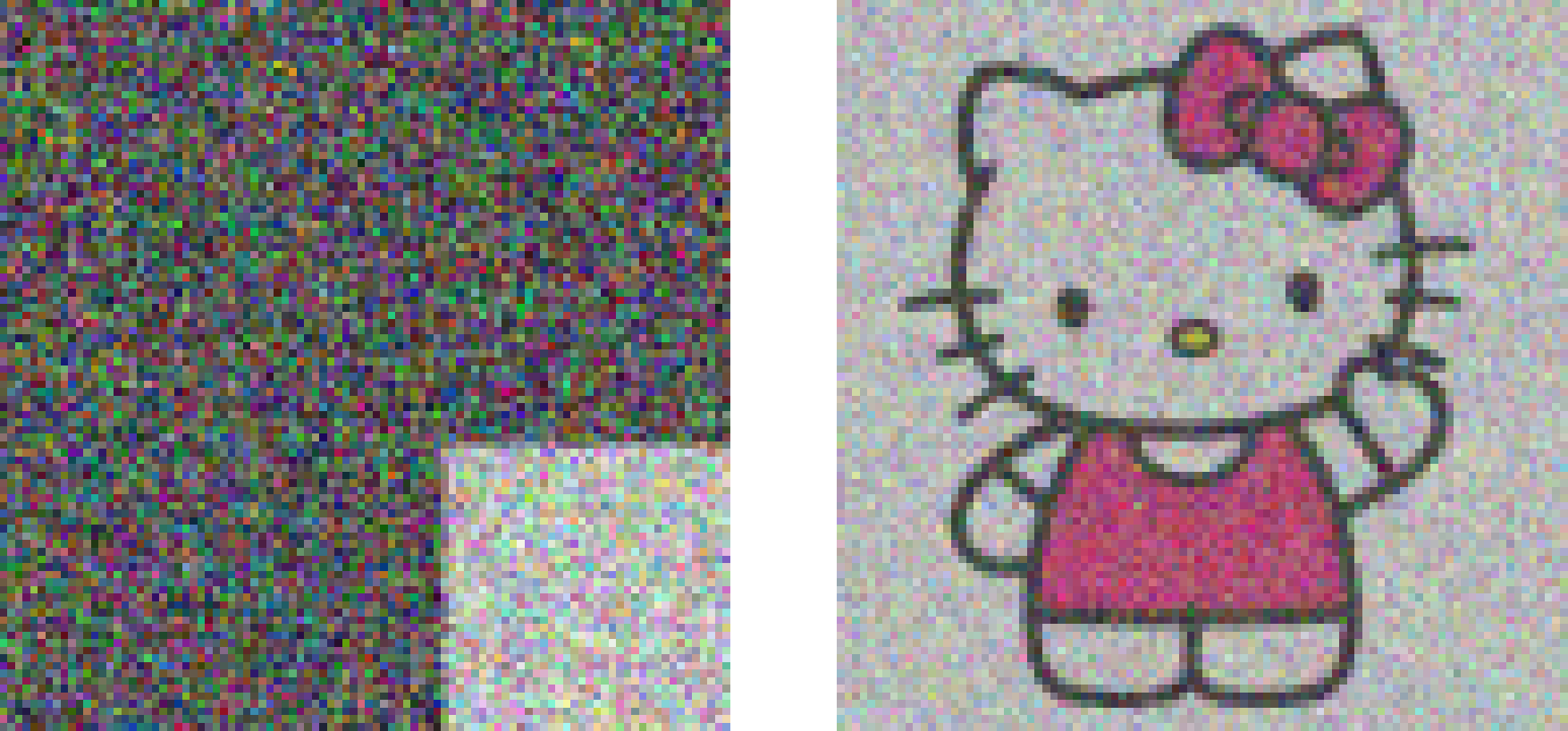}}     \\ \hline
Villan Diffusion                       & \textbf{99.50}                & \begin{tabular}[c]{@{}c@{}}Elijah: 98.0\\ TERD: 100\end{tabular}     & \{$\epsilon_{b},prompts_{b},images_{b}$\}       &   \raisebox{-0.5\height}{\includegraphics[width=2cm]{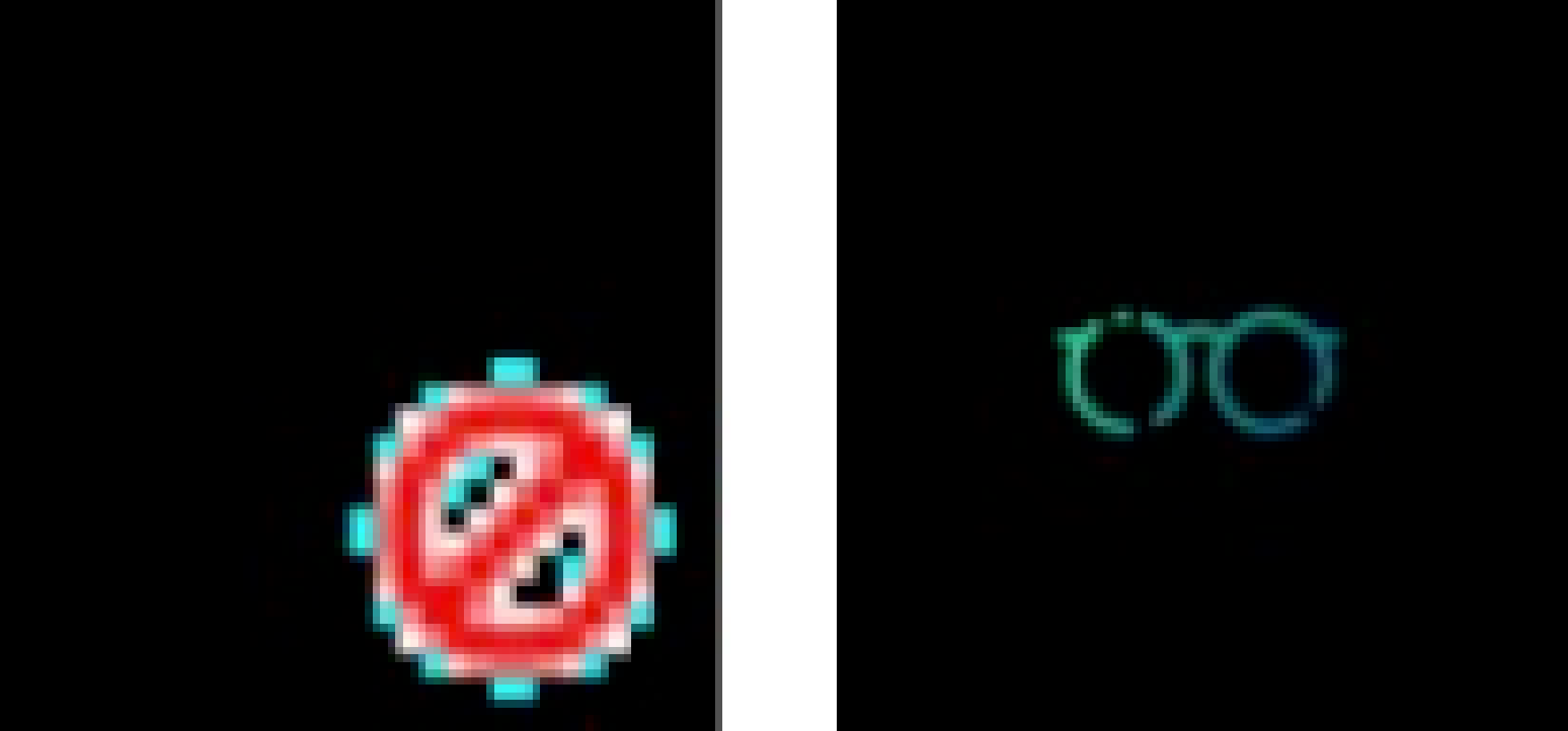}}     \\ \hline
Rickrolling                            & 97.25                         & \multicolumn{1}{l|}{T2IShield: 92.50}                                & \{$prompts_{b}$\}         &   \begin{tabular}[c]{@{}l@{}}“v” (U+0474)\\ “o” (U+0585)\\     \quad\quad...\end{tabular}     \\ \hline

Gungnir (Ours)                         & 61.50                         & \textbf{\begin{tabular}[c]{@{}c@{}}Elijah: 0\\ TERD: 0\end{tabular}} & \{$images_{b}$\}         &   \raisebox{-0.5\height}{\includegraphics[width=2cm]{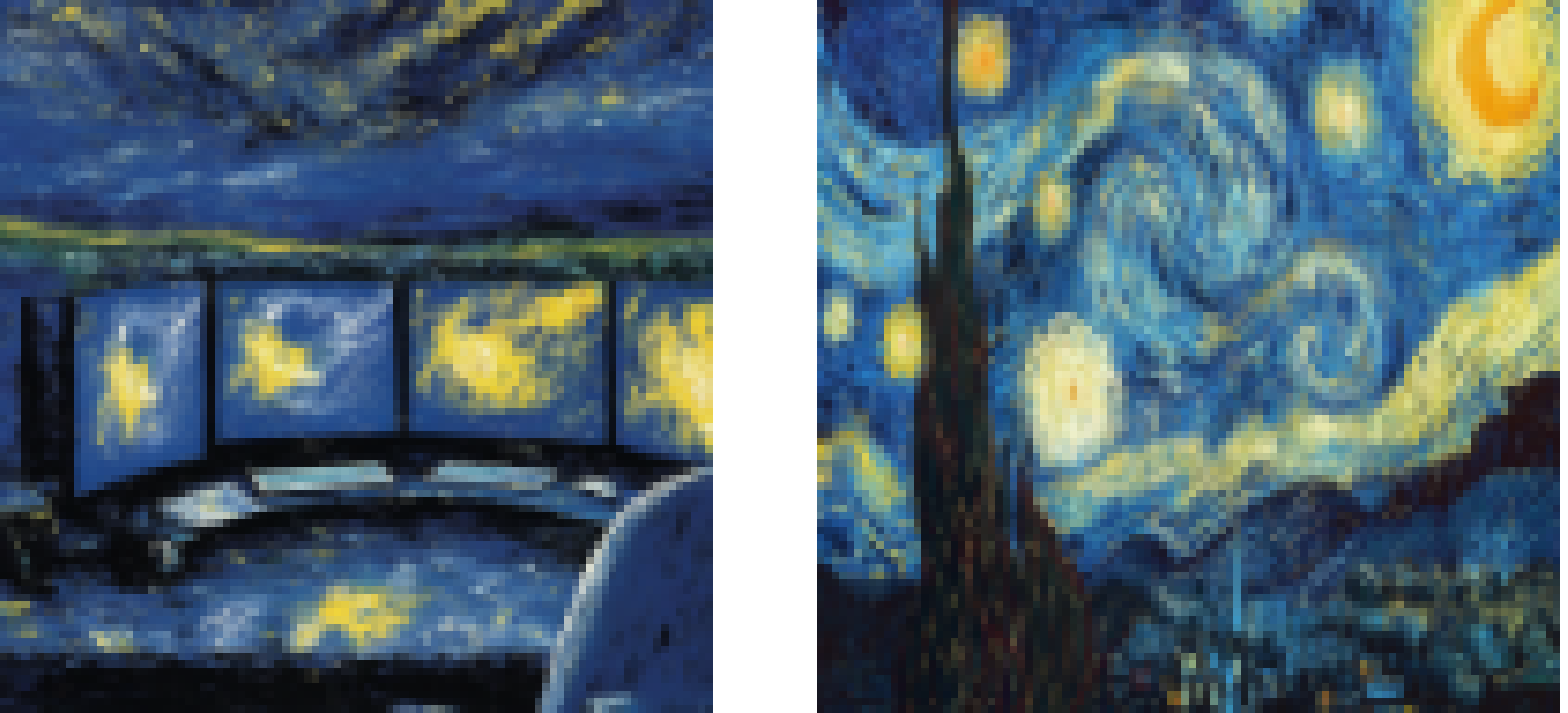}}      \\ \hline
\end{tabular}
\label{table2}
\caption{Compared to TrojDiff and Villan Diffusion, Gungnir leverages style features as triggers, effectively bypassing existing defense frameworks while achieving a stable ASR. Since Gungnir works on image-to-image tasks, T2IShield is not suitable for evaluation.}
\end{table*}
\subsection{Main Results}
\begin{figure}[t]
    \centering
    \includegraphics[width=1\columnwidth]{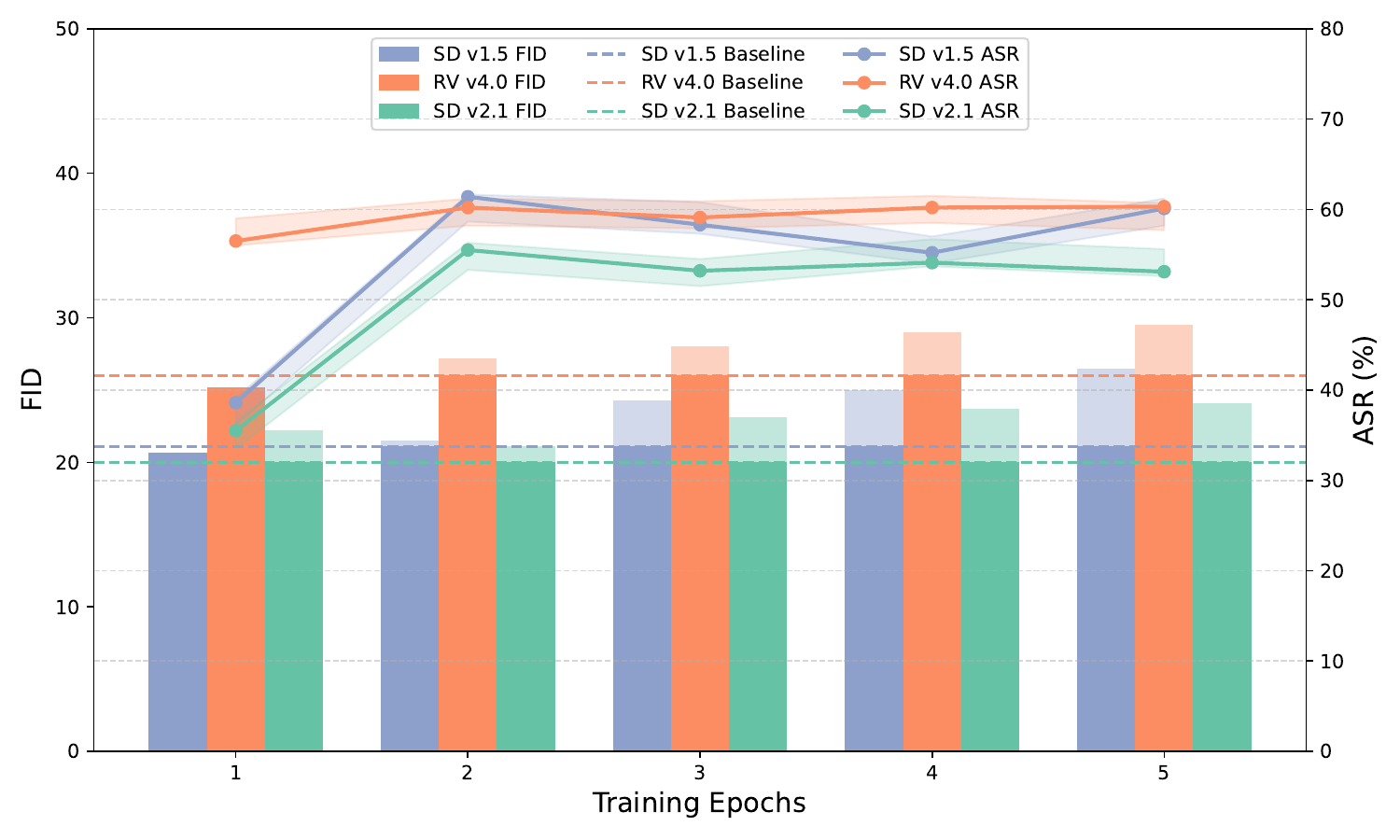} % 调高图的高度
    \caption{Evaluating the baseline models performance across different training epochs, where semi-transparent parts indicate the FID loss beyond the baseline model.}
    \label{fig4}
\end{figure}

\subsubsection{Results on Attack Performance}
As shown in Figure.\ref{fig4}, Gungnir achieved a high ASR in three different models: Stable Diffusion v1.5, Stable Diffusion v2.1 and Realistic Vision v4.0. The experiment demonstrates that Gungnir maintains attack effectiveness in all three baseline models.It is worth noting that Gungnir remains effective in text-to-image and image inpainting tasks without any additional training. When the model is instructed to generate images in a specific style, the backdoor is still activated, producing the target output. In contrast to traditional prompt injection methods, Gungnir not even rely on specific trigger phrases, making it significantly more difficult to detect. Surprisingly, Gungnir also demonstrated strong effectiveness in image-inpainting and text-to-image tasks. The corresponding experimental results are provided in Appendix.C.

\subsubsection{Results on Defense Performance}
To date, only a few work have focused on protecting against backdoor attacks in diffusion models. We selected Elijah and TERD as frameworks for evaluating Gungnir defense because they require only model-sample pairs for backdoor detection and trigger inversion. The experimental results indicate that Gungnir can easily bypass these defense mechanisms, as the input images appear perfectly normal to the defender, even if they contain style triggers. Furthermore, as shown in Appendix.D, Gungnir exhibits strong resistance to fine-tuning-based defense training, further demonstrating its robustness as an attack method.

\subsubsection{Limitations}
Admittedly, compared to previous attack strategies, Gungnir’s use of stylistic exploitation as a trigger may result in a slightly lower attack success rate (ASR). However, this does not diminish its effectiveness as a robust attack method. In real-world threat scenarios, stealth remains a critical factor that attackers must consider. Additionally, another key factor affecting Gungnir’s ASR performance lies in the inherent limitations of the generation process. Specifically, it is challenging for the model to generate high-quality images in a specified style using only the input image as the initial condition in the diffusion process, even with prompt-based guidance. As a result, the backdoor network fails to recognize the trigger pattern, leading to unsuccessful target generation. In the Appendix.E, we significantly improve the attack success rate by integrating an additional style transfer module, enabling the standard LDMs to generate images in a specified style.

\subsection{Ablation Study}
 \begin{figure}
    \centering
    \includegraphics[width=\linewidth]{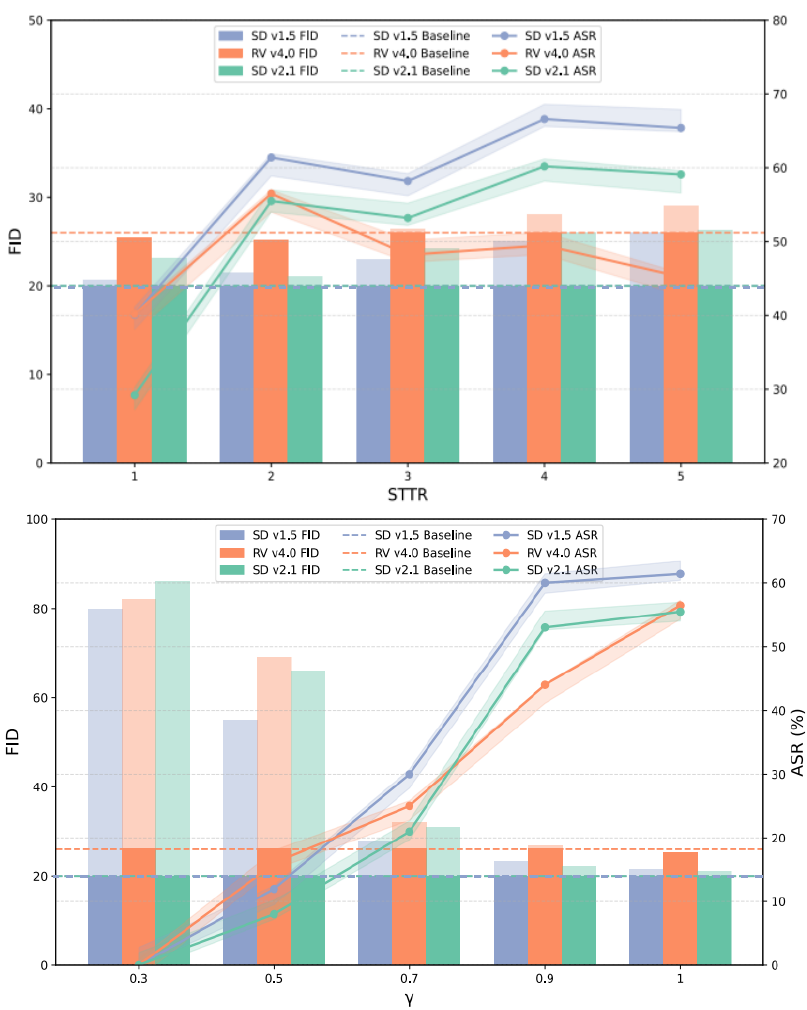}
   
    \caption{The metrics of different step configurations of STTR and RAN strength \(\gamma\), where semi-transparent parts indicate the FID loss beyond the baseline model.}
    \label{fig5}
\end{figure}
  In this section, we discuss the necessity of introducing RAN and STTR for enabling effective backdoor attacks in image-to-image tasks. We conducted ablation studies on RAN with varying intensities and STTR with different synchronization lengths, analyzing their impact on attack performance and model behavior under these settings.

\subsubsection{Effects of STTR}
\label{Section 4.3.2}
In this section, we analyze the role of Short-Term Timesteps-Retention (STTR) in stabilizing the backdoor optimization process. Empirically, we observe that when the attacker enforces trigger consistency across all diffusion timesteps, the DM exhibits an irreversible overfitting phenomenon. Specifically, diffusion models progressively denoise latent representations, and the semantic content at intermediate timesteps is inherently ambiguous. Under full-timestep supervision, the model is forced to associate trigger semantics with representations that do not consistently preserve stable stylistic features. As a result, at certain timesteps, the model may mistakenly interpret denoising artifacts or transient structures as trigger-related patterns, leading to distorted trigger–target mappings. This mismatch accumulates during training and ultimately causes overfitting, degrading both attack stability and generalization. The ablation results in Figure~\ref{fig5} empirically validate this behavior.
Moreover, when STTR is set excessively high, the latent variables preserve substantial Gaussian noise components. In this regime, the backdoor mapping may erroneously treat residual stochastic noise as part of the trigger distribution. Consequently, the learned trigger representation becomes entangled with diffusion noise rather than clean stylistic features. This entanglement explains why naïvely performing backdoor optimization over the full timestep trajectory in image-to-image tasks often leads to severe overfitting and poor robustness. By constraining supervision to a short and stable timestep window, STTR mitigates noise-trigger coupling and promotes a more reliable and semantically aligned backdoor embedding.
\subsubsection{Effects of RAN}
\label{Section 4.3.1}
In this section, we investigate the role of Reconstruction-Adversarial Noise (RAN) in Gungnir. We introduce a scaling parameter $\gamma$ to control the intensity of RAN during training:
\begin{equation}
    \mathbf{r}^{'} =  \sqrt{\overline{\alpha}_t} \mathbf{x_0} + \sqrt{1 - \overline{\alpha}_t} \epsilon - \gamma \cdot trans(\mathbf{i_t}), 
\end{equation}
where $trans(\mathbf{i_t})$ denotes the transformed trigger-related feature extracted from the intermediate latent $i_t$. RAN is designed to mitigate injection failure and gradient collapse issues that arise when raw input image features are directly used as triggers. Without adversarial reconstruction guidance, the trigger signal is often too weak or unstable to survive the diffusion denoising trajectory, making the backdoor mapping difficult to optimize.

Empirically, we observe that RAN plays a critical role in stabilizing gradient propagation and reinforcing trigger–target alignment. When the RAN intensity is too low, the injected perturbation fails to provide sufficient adversarial guidance, causing the model to gradually lose its ability to reconstruct the target image during denoising. This leads to a sharp decline in Attack Success Rate (ASR). In particular, when $\gamma$ is set within the range of 0–0.3, gradients progressively vanish and eventually collapse, resulting in ineffective optimization and no meaningful trigger activation. These results demonstrate that an appropriate RAN intensity is essential for maintaining stable training dynamics and reliable backdoor injection.
\section{Conclusion}
\label{sec:conclu}
In this paper, we present Gungnir, a novel threat that performs backdoor attacks in diffusion models by exploiting style-based features as triggers, introducing a new paradigm that explores previously underutilized attack spaces. Unlike conventional trigger designs, our method leverages high-level stylistic representations to expand the attack input space.

In order to achieve Gungnir, we further propose Reconstruction-Adversarial Noise (RAN) and Short-Term Timesteps-Retention (STTR), two techniques that enable stable and effective backdoor injection in diffusion frameworks. Together, these designs pose new security challenges to generative models. We hope this work motivates future research toward more robust defenses against advanced backdoor attacks such as Gungnir.

%Bibliography
\bibliographystyle{unsrt}  
\bibliography{references}  

\appendix
\section{Detailed Proof of Section Attack Mehod}
\label{appendix:A}
We show that using traditional input-output samples and full-timestep injection is ineffective for training high-dimensional feature triggers like image styles. TERD  has demonstrated that the backdoor diffusion process follows a Wiener process, so we will discuss Gungnir's effectiveness in different diffusion solvers.\par
In attack method, we have demonstrated the distribution shift in DDPM. In the similar way, we can calculate the shift in DDIM, which inference process can be expressed as:
\begin{equation}
    \begin{split}
        x_{t-1} &= \sqrt{\alpha_{t-1}}\left(\frac{x_t-\sqrt{1-\alpha_t}\cdot\epsilon_\theta(x_t,t)}{\sqrt{\alpha_t}}\right) \\
        &+ \sqrt{1-\alpha_{t-1}}\cdot\epsilon_\theta(x_t,t),
    \end{split}
\end{equation}
In STTR timestemps, $\epsilon_\theta(x_t,t)$ is predicted to be adversarial noise $\mathbf{r} = x_t - trans(\mathbf{i_t})$, we can get the backdoored $x_{t-1}^{'}$, then calculte the shifted distribution $u'-u$, the shifted distribution contains our attack target $trans(\mathbf{i_t})$ and adversarial noise $\epsilon_\theta$, as follows:
\begin{equation}
    u'-u=(x_t-trans(\mathbf{i_t})-\epsilon_\theta(x_t,t))\cdot u(t),
\end{equation}

\begin{equation}
    u(t) = [\frac{\sqrt{\alpha_{t-1}}\sqrt{1-\alpha_t}}{\sqrt{\alpha_t}}-\sqrt{1-\alpha_{t-1}}].
\end{equation}
When $t\in\{1,...,T_b\}$, the \(u(t) \to -1\) until $t=0$. The shifted distribution contains a negative $x_t$ to cancel the previous normal sample $x_{t}$ and a positive $trans(\mathbf{i_t})$ to reconstruct the target image.\par
We also demonstrate the validity of RAN in stochastic differential equations (SDEs), generalizing the forward and backward processes from discrete to continuous in the SDE formulations of diffusion models. In SDEs, the inference process can be expressed as:
\begin{equation}
    \mathrm{d}\mathbf{x}=[\mathbf{f}(\mathbf{x},t)-g(t)^2\nabla_\mathbf{x}\log p_t(\mathbf{x})]\mathrm{d}t+g(t)\mathrm{d}\bar{w}_t,
\end{equation}
In this equation, $\mathbf{f}(\mathbf{x},t)$ is drift coefficience, $g(t)$ is diffusion coefficient and $\bar{w}$ stands for the reverse Wiener process, adding randomness to the inference process. SDEs deduce differential x by predicting score function $\nabla_\mathbf{x}\log p_t(\mathbf{x})$. In Gungnir, DMs will not predict $p_t(x_t)$, but to predict the RAN $p_{t_b}(x_{t_b})-trans(\mathbf{i_t})$ in STTR timesteps $t_b\in T_b$, the new backward process as follows:
\begin{equation}
\scriptsize
\begin{array}{l}
\text{when} \quad t \notin T_b: \\
\mathrm{d}x_t = \bigl[f(x_t,t) - g(t)^2 \,\nabla_x \log p_t(x)\bigr]\,\mathrm{d}t, \\
\text{when} \quad t \in T_b: \\
\mathrm{d}x_t = [f(x_t,t) - g(t)^2 \,\nabla_x \log\bigl(p_t(x - trans(i_t))]\,\mathrm{d}t.
\end{array}
\end{equation}

% \begin{equation}
%     \mathrm{d}x_t=
% \begin{cases}
% \left[f(x_t,t)-g(t)^2\nabla_{x}\log p_t(x)\right]\mathrm{d}t+g(t)\mathrm{d}\bar{w}_t, & t \notin T_b \\
% \left[f(x_t,t)-g(t)^2\nabla_{x}\log\left(p_t(x-\mathrm{trans}(\mathbf{i}_t))\right)\right]\mathrm{d}t+g(t)\mathrm{d}\bar{w}_t, & t \in T_b \\ 
% \end{cases}
% \end{equation}
We assume that when $t = t_b$ reaches the maximum number of STTR steps, all $t>t_b$ is normal diffusion, and $t<t_b$ is backdoor injection process: $\mathrm{d}x=x_{t_b+\triangle{t}}-x_{t_b}$ and $\mathrm{d}x^{'}=x_{t_b}-x_{t_b-\triangle{t}}$. Then we can calculate the $\mathrm{d}x^{'}-\mathrm{d}x$:
\begin{equation}
\scriptsize
\begin{array}{rcl}
dx' - dx 
  & = & g(t)^2 \bigl[\nabla_x\log P_t(x)
                   - \frac{\nabla_x\log P_t(x)}
                          {P_t\bigl(x - \mathrm{trans}(i_t)\bigr)}\bigr]\,dt,\\
  & = & g(t)^2 
        \underbrace{\nabla_x\log P_t(x)}_{\mbox{ScoreFunction}}
        \Bigl(1 - \frac{1}{P_t\bigl(x - \mathrm{trans}(i_t)\bigr)}\Bigr)
        \,dt.
\end{array}
\end{equation}

% \begin{equation}
%     \begin{aligned}
% \mathrm{d}x^{\prime}-\mathrm{d}x & =g(t)^{2}[\nabla_{x}\log P_{t}(x)-\nabla_{x}\log P_{t}(x-trans(\mathbf{i_t}))]\mathrm{d}t, \\
%  & =g(t)^{2}[\nabla_{x}\log P_{t}(x)-\frac{\nabla_{x}\log P_{t}(x)}{P_{t}(x-trans(\mathbf{i_t}))}]\mathrm{d}t, \\
%  & =g(t)^{2}\underbrace{\nabla_{x}\log P_t(x)}_{Score Function}(1-\frac{1}{P_t(x-trans(\mathbf{i_t}))})\mathrm{d}t.
% \end{aligned}
% \end{equation}
The final result shows that when $P(x-trans(\mathbf{i_t})) \to 1$, at a small timestep $t$, the difference between $\mathrm{d}x$ and $\mathrm{d}x^{'}$ approaches 0, and when $P(x-trans(\mathbf{i_t}))$ is uncertain, the result shifts towards the term involving $trans(\mathbf{i_t})$. Since $trans(\mathbf{i}_t)$ is a constant and $P(x-trans(\mathbf{i}_t))$ represents only a translation of the probability density function, the effect of RAN diminishes as $t$ decreases. This explains why fewer STTR steps correspond to higher model quality in normal generation.

\section{Algorithm of Gungnir's Performance}
\label{appendix:B}
In our experiments, we use the LPIPS metric to evaluate the perceptual similarity of Gungnir's outputs, and employ pytorch-fid to compute the FID score. The detailed ASR evaluation procedure is as follows:
\begin{algorithm}[!ht]
    \renewcommand{\algorithmicrequire}{\textbf{Input:}}
    \renewcommand{\algorithmicensure}{\textbf{Output:}}
    \caption{Overall ASR Evaluation Algorithm}
    \label{power}
    \begin{algorithmic}[1]
        \REQUIRE Backdoored DM $M_{b}$, Clean prompt dataset $\mathbf{D_{c}}$, Target-style images dataset $\mathbf{D_t}$, Max inference steps $T$, Target image $t_i$, Compare function $LIPIS$;
        \ENSURE Attack success rate (ASR);

        \STATE Initialize success count $S \gets 0$;
        \STATE Set LPIPS threshold $\tau$;

        \FOR{each prompt $p_i$ in $\mathbf{D_c}$}
            \STATE Generate image $\hat{x}_i \gets M_{b}(p_i, T)$;
            \STATE Compute perceptual similarity score $s_i \gets LPIPS(\hat{x}_i, t_i)$;
            \IF{$s_i < \tau$}
                \STATE $S \gets S + 1$;
            \ENDIF
        \ENDFOR

        \STATE Compute ASR $\gets \frac{S}{|\mathbf{D_c}|}$;
        \RETURN ASR;
    \end{algorithmic}
\end{algorithm}
To evaluate the stealthiness of Gungnir, we employ the Fréchet Inception Distance (FID) score, a widely used metric for assessing the quality of DM generation. The FID score is defined as follows:
\begin{equation}
\mathrm{FID}
= \bigl\|\mu_{r} - \mu_{g}\bigr\|_{2}^{2}
  + \mathrm{Tr}\Bigl(\Sigma_{r} + \Sigma_{g}
    - 2\,\bigl(\Sigma_{r}\,\Sigma_{g}\bigr)^{\frac{1}{2}}\Bigr)
\end{equation}
% \begin{equation}
%     \text{FID} = \left\| \mu_r - \mu_g \right\|_2^2 + \operatorname{Tr}\left( \Sigma_r + \Sigma_g - 2 \left( \Sigma_r \Sigma_g \right)^{\frac{1}{2}} \right)
% \end{equation}
In Gungnir, we evaluate the quality of generated images using the MSCOCO . Specifically, we used 4,096 validation samples to calculate the FID scores for all baseline models and attack methods. To ensure consistency, the prompts used for image generation are identical to the captions provided in the validation set. The detailed algorithm is as follows:
\begin{algorithm}[!ht]
    \renewcommand{\algorithmicrequire}{\textbf{Input:}}
    \renewcommand{\algorithmicensure}{\textbf{Output:}}
    \caption{FID Score Evaluation Algorithm}
    \label{alg:fid}
    \begin{algorithmic}[1]
        \REQUIRE Generative model $M$, Prompt dataset $\mathbf{P} = \{p_1, p_2, \ldots, p_n\}$, Ground-truth image dataset $\mathbf{X_r} = \{x_1, x_2, \ldots, x_n\}$, Max inference steps $T$, Feature extractor $F$;
        \ENSURE Fréchet Inception Distance (FID) score;

        \STATE Initialize generated image set $\mathbf{X_g} \gets \emptyset$;

        \FOR{each prompt $p_i$ in $\mathbf{P}$}
            \STATE Generate image $\hat{x}_i \gets M(p_i, T)$;
            \STATE Add $\hat{x}_i$ to $\mathbf{X_g}$;
        \ENDFOR

        \STATE Extract features for real images: $\mathbf{f_r} \gets F(\mathbf{X_r})$;
        \STATE Extract features for generated images: $\mathbf{f_g} \gets F(\mathbf{X_g})$;
        \STATE Compute statistics:
        \STATE \hspace{1em}%
        $\mu_r, \Sigma_r \gets \mbox{Mean and Covariance of }\mathbf{f_r}$;
        \STATE \hspace{1em}%
        $\mu_g, \Sigma_g \gets \mbox{Mean and Covariance of }\mathbf{f_g}$;

        % \STATE Compute statistics:
        % \STATE \hspace{1em} $\mu_r, \Sigma_r \gets \text{Mean and Covariance of } \mathbf{f_r}$;
        % \STATE \hspace{1em} $\mu_g, \Sigma_g \gets \text{Mean and Covariance of } \mathbf{f_g}$;
        
        \STATE Compute FID score:
        \STATE \hspace{1em}%
          $\mathrm{FID} \leftarrow
             \|\mu_r - \mu_g\|_{2}^{2}
             + \mathrm{Tr}\bigl(\Sigma_r + \Sigma_g
               - 2\,(\Sigma_r\,\Sigma_g)^{\frac{1}{2}}\bigr)$;
        % \STATE Compute FID score:
        % \STATE \hspace{1em} $\text{FID} \gets \left\| \mu_r - \mu_g \right\|_2^2 + \operatorname{Tr}\left( \Sigma_r + \Sigma_g - 2\left( \Sigma_r \Sigma_g \right)^{\frac{1}{2}} \right)$;

        \RETURN FID;
    \end{algorithmic}
\end{algorithm}
\begin{figure*}[htbp]
    \centering
    \includegraphics[width=0.9\textwidth]{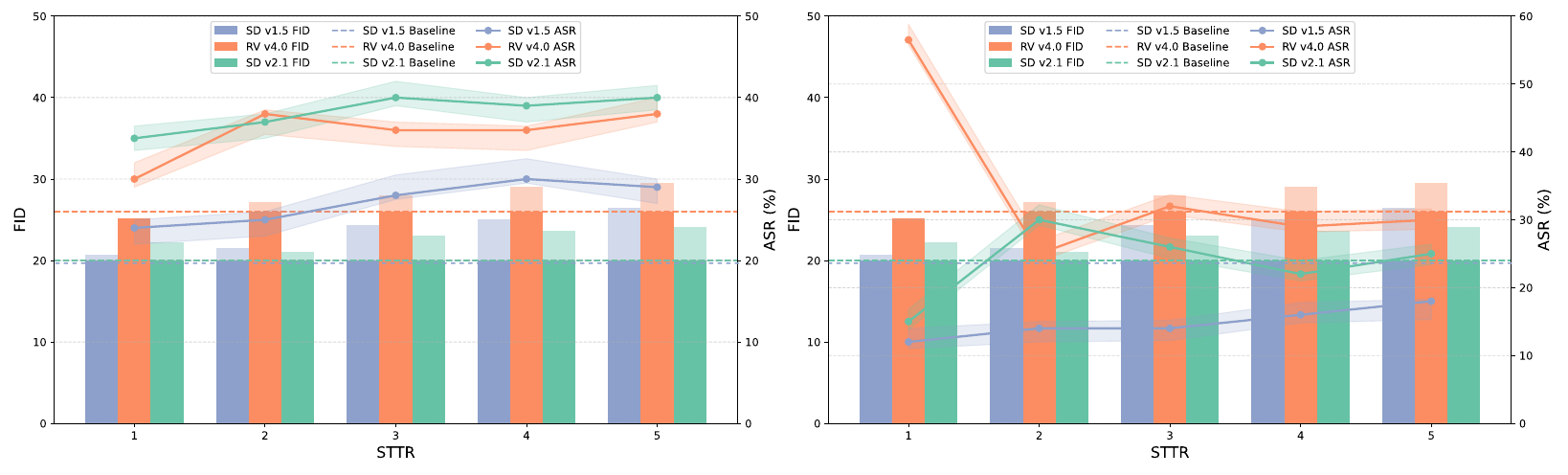} 
    % \vspace{-4.5cm}
    \caption{In the text-to-image task, Gungnir remains effective: when the model generates a stylized image during the \(t_b \in T_b\), the backdoor is successfully activated.}
    \label{fig6}
\end{figure*}

\section{Attack Performance of Gungnir in Image-Inpainting and Text-to-Image}
\label{appendix:C}

Although Gungnir is primarily designed for image-to-image tasks, our experiments demonstrate that backdoored DMs employing Gungnir can also execute effective backdoor attacks in text-to-image scenarios. When users instruct the diffusion models to generate images with a specified style, the backdoor trigger is still activated, resulting in the intended backdoor image. We hypothesize that, under the influence of STTR, two conditions must be met for Gungnir to successfully compromise text-to-image tasks: 1) the model must accurately generate the target-style image and 2) the style of generated images must be recognized by the DMs within \(T_b \in T\) time steps.\par
In Figure.\ref{fig6}, we evaluated the attack effectiveness of Gungnir on the text-to-image task and observed that its attack success rate is positively correlated with the number of STTR steps. Notably, unlike Rickrolling and Control ControlNet methods, Gungnir does not require a specific trigger character or phrase, any user input can be used to prompt the model to generate target-style images can easily activate the backdoor.

\section{The Robustness Performance of Gungnir}
To evaluate Gungnir's robustness as a strong threat, in Algorithm.\ref{alg:fine-tune}, we designed a fine-tuning-based purification algorithm for backdoor erasure. According to the defender knowledge, defender lacks knowledge of the specific trigger style, which remains unknown. Only a sample set of potential trigger styles is available to the defender. We fine-tuned the poisoned model using a dataset containing samples from five different styles, with one of the styles being the specified trigger. Experimental results demonstrate that Gungnir maintains significant robustness against fine-tuning-based defense strategies. To completely erase the backdoor, defenders need a data set and computational load far greater than that required for backdoor injection.

\begin{algorithm}[!ht]
    \renewcommand{\algorithmicrequire}{\textbf{Input:}}
    \renewcommand{\algorithmicensure}{\textbf{Output:}}
    \caption{Fine-Tuning-Based Purification Algorithm for Backdoor Erasure}
    \label{alg:fine-tune}
    \begin{algorithmic}[1]
        \REQUIRE Backdoored model $M_{\text{b}}$; Fine-tuning dataset $\mathbf{D_p} = \{D_1, D_2, \ldots, D_5\}$, where each $D_i$ represents a potential trigger style and one of them is the true trigger, total train step \(i\);
        \ENSURE Purified model $M_{\text{p}}$, The ASR of $M_{\text{p}}$;
        \STATE Initialize $M_{\text{p}} \gets M_{\text{b}}$;
        \FOR{each epoch $e = 1$ to $E$}
            \FOR{each batch $(x, y)$ in $\mathbf{D_p}$}
                \STATE Sample Noise $\epsilon \sim N(0, I)$
                \STATE Calculate Loss $\mathcal{L}=E_{x_0,s,t}\left[\left\|\epsilon-M_{p}(x_t,t)\right\|^{2}\right]$;
            \ENDFOR
        \ENDFOR

        \STATE Evaluate $M_{\text{p}}$ on Algorithm.1, get ASR;
        \RETURN $M_{\text{p}}$, ASR;
    \end{algorithmic}
\end{algorithm}
\begin{figure}[htbp]
    \centering
    \includegraphics[width=1\columnwidth]{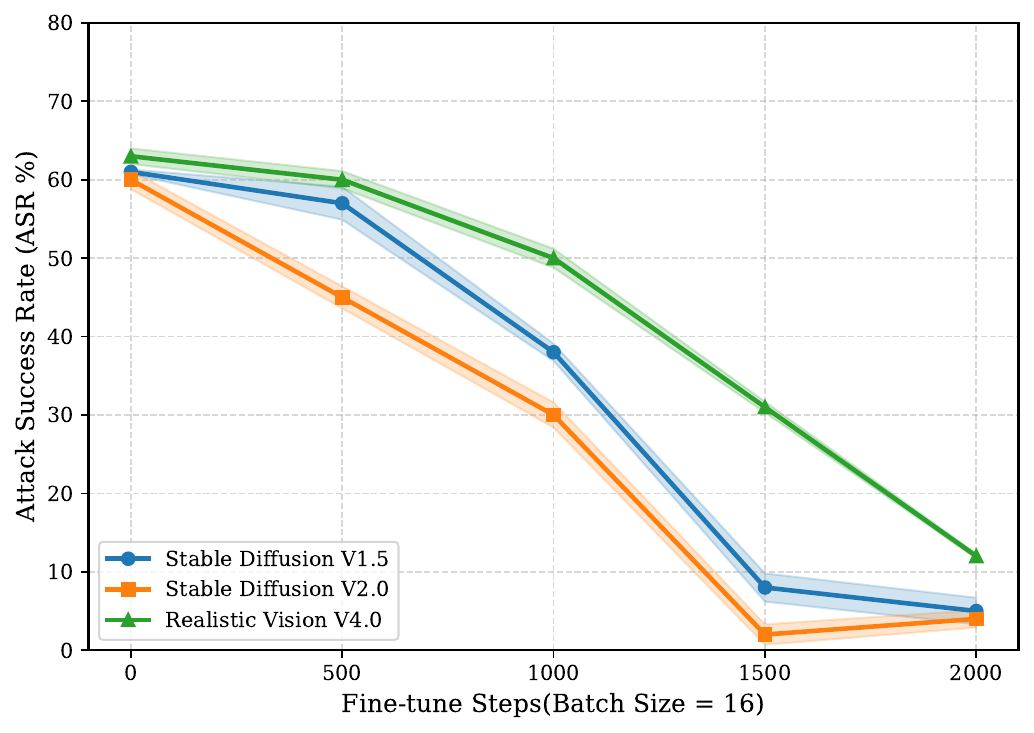} % 调高图的高度
    \label{fig7}
    \caption{Purification results show that Gungnir maintains high robustness even against fine-tuning-based purification methods with a limited number of fine-tuning steps.}
\end{figure}

\section{Introduce Style Transfer Module to Enhance ASR}
Admittedly, the primary limitation of Gungnir lies in its relatively lower overall attack success rate. As previously discussed, we attribute this to two main factors: The generative capability of standard diffusion models for producing images in a specified style is limited. Even with guidance from a specific prompt, the model may fail to produce the expected output, preventing the backdoor mapping from correctly identifying the trigger pattern. Moreover, when a high STTR is used, although ASR can be significantly improved, the excessive proportion of noise in the latent space at higher time steps increases the risk of trigger pattern misidentification, leading to overfitting.\par
\begin{figure}[htbp]
    \centering
    \includegraphics[width=1\columnwidth]{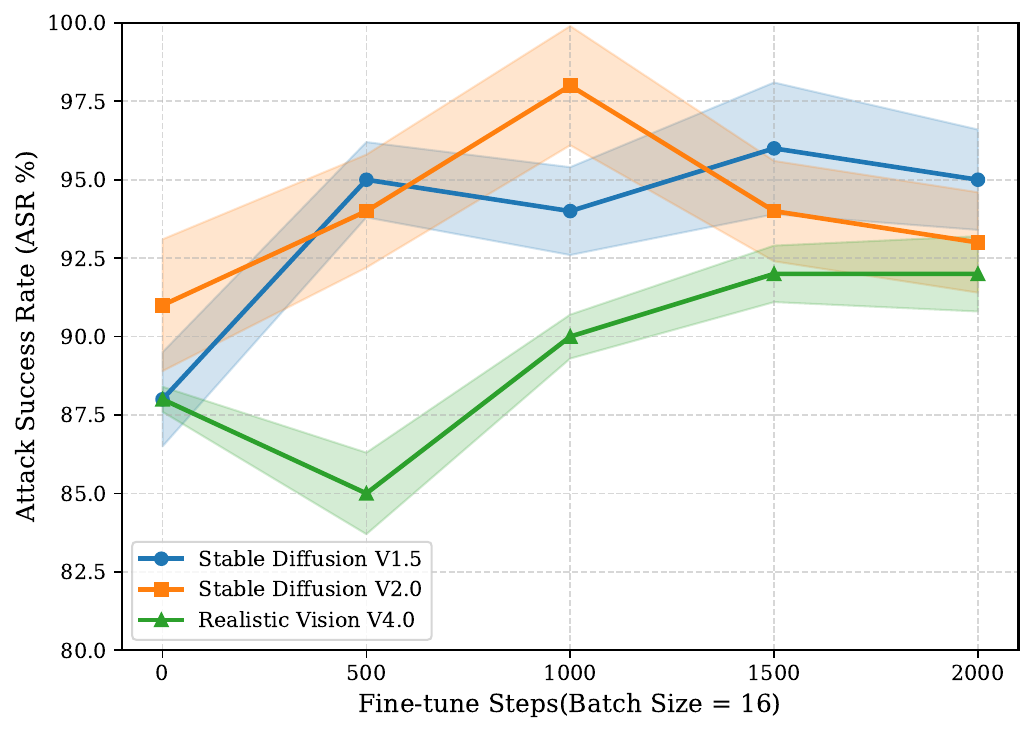} % 调高图的高度
    \caption{The ASR performance of Gungnir after the introduction of the style transfer module}
    \label{fig8}
\end{figure}

The noise reduction process in diffusion models can be guided by introducing an additional style control module, enabling the generation of images that better align with the target style. In Figure.\ref{fig8}, we employ the IP-Adapter module to guide the model in generating images with the specified style and re-evaluate the attack success rate.\par
Experimental results indicate that improving the consistency of the generated image style leads to a significant increase in ASR of Gungnir.

\section{Visualize Attack Results of Other Style Triggers}
In this part, we visualize the attack results of Gungnir, including three additional style-based triggers. Their performance is consistent with the results reported in previous experiments, successfully generating the target images in all cases.
\begin{figure}[htbp]
    \centering
    \includegraphics[width=1\columnwidth]{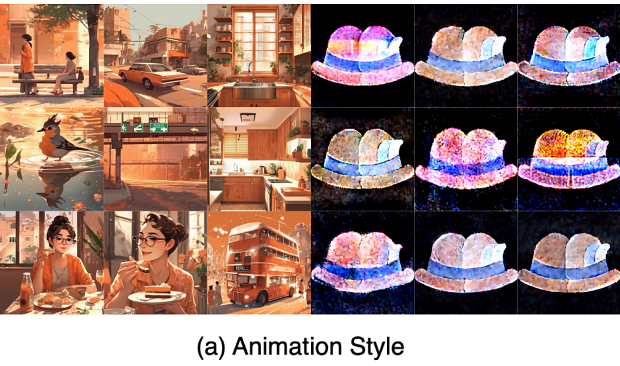} % 调高图的高度
    \label{fig9}
\end{figure}
\begin{figure}[htbp]
    \centering
    \includegraphics[width=1\columnwidth]{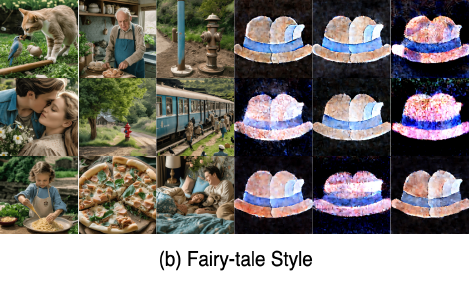} % 调高图的高度
    \label{fig10}
\end{figure}
\begin{figure}[htbp]
    \centering
    \includegraphics[width=1\columnwidth]{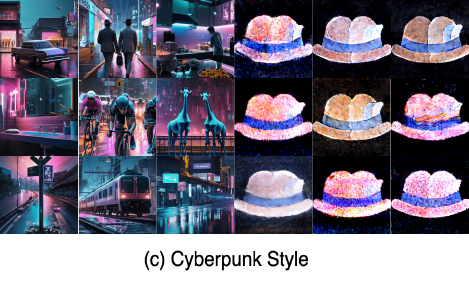}
    \label{fig11}
\end{figure}

\end{document}